\documentclass[11pt]{article}

\usepackage{cmap}
\usepackage[utf8]{inputenc}
\usepackage[T2A,T1]{fontenc}
\usepackage{lmodern}
\usepackage[russian,english]{babel}
\usepackage[margin=1in]{geometry}
\usepackage{booktabs}
\usepackage{amsmath}
\usepackage{graphicx}
\usepackage{url}
\usepackage[hidelinks]{hyperref}
\usepackage{orcidlink}
\usepackage{natbib}

\newcommand{\ru}[1]{\foreignlanguage{russian}{#1}}

\title{Label-Wording Augmentation in a Typed-Decision Encoder:\\
Gains on Unfamiliar Words,\\
Not on Abbreviations or Codes}

\author{Vladimir Beskorovainyi\,\orcidlink{0009-0004-7005-6242}\\
Besk Tech, Moscow, Russia\\
\texttt{https://besk.tech}}

\date{November 2026}

\begin{document}
\maketitle

\begin{abstract}
Typed-decision models score options that the caller supplies at request time,
so their accuracy depends on label wording that the model's developer does not
control. We compare two releases of a 307M-parameter multilingual
cross-encoder, Ru-Jev, trained on thirteen Russian-language tasks; the later
release adds an augmentation that presents labels in several surface forms.
Holding questions fixed and varying only how eighteen routing labels are
worded, we evaluate ten label vocabularies, produced by three separate
processes and checked for overlap with the training data, with five
candidates per question and, for eight of them, also with eighteen. On held-out vocabularies made of ordinary or
borrowed words (English-derived jargon, office slang, department names), the
augmented release has higher point accuracy in all fourteen cells, by up to
0.24; eleven of these differences survive a Holm correction at the question
level (slang with eighteen candidates and descriptive sentences with five do
not), and eight have a label-level interval that excludes zero; the slices of
one vocabulary are nested, so these counts are not independent. Two runs
trained by mistake without the transliterated label forms stay at the earlier
release's level on jargon, which is consistent with the jargon gain coming from
transliteration. The comparison is between two trained releases, not a
controlled ablation. On abbreviations and opaque codes the augmented release
gives no usable gain: on the 17 abbreviations unrelated to training it is lower
as released ($-0.060$), but one label carries 52 of the 58 questions and one
bias per option, fitted out of fold on these questions, removes the difference
(0.000);
on opaque codes its gain with eighteen candidates leaves it near chance (0.061
against 0.056) once labels related to training options are removed and does
not survive the same correction. Several word-vocabulary gains also rest partly
on one label that the earlier release almost never chose; the jargon gain
largely survives that correction. In most cells where accuracy rises, the
risk--coverage area falls, but with eighteen candidates confident errors rise
on jargon and slang. On two datasets neither
release saw, accuracy moves by $-8.4$ and $+5.6$ points; one additive bias per
option, fitted on half of the rows, narrows these to $-3.4$ and $+1.4$ points,
the headline gap stays distinguishable from zero over questions, and neither
difference excludes zero once labels are resampled. A second training run of each recipe, the one for the augmented
recipe trained under a reading rule recorded beforehand, puts the gap between the recipes on jargon far above
seed noise, and on a third unseen dataset, scientific rubrics, only just above
it; rankings of training checkpoints disagree across the three unseen datasets.
We also report four evaluation hazards from comparing the two releases.
\end{abstract}

\section{Introduction}

Text classifiers have historically fixed their label set at training time. A
growing class of interfaces does not: the caller supplies the candidate options
at request time, in the caller's own words. This covers zero-shot
classification by entailment \citep{yin2019}, prompt-based classification with
verbalizers \citep{schick2021}, structured output from instruction-following
models, and compact \emph{typed-decision} models that return a decision of a
declared type with a probability and generate no free text
\citep{deusser2026jev, wang2026chinesejev, laya}.

When the label set arrives at request time, accuracy becomes a function of the
strings the caller chose. Benchmarks usually score models on the label names
the dataset authors wrote, while deployments often use names the organisation
already has: department names, internal codes, abbreviations and borrowings
from English. That label wording affects behaviour is established.
\citet{webson2022} find prompt-based models far more sensitive to the choice of
target words than to the instruction; \citet{gao2021} search over label words
because the choice matters; \citet{holtzman2021} identify surface form
competition in multiple-choice scoring; \citet{gao2023labeldesc} show that
training on label descriptions improves robustness to verbalizer choice. In
vision, robustness gained from one family of corruptions often fails to
transfer to others \citep{geirhos2018}.

We ask a practical question. If training presents each label in several
surface forms, which kinds of unfamiliar label wording does the model then
handle better? We answer it with probes that hold the questions fixed and vary
only the wording, at two candidate-set sizes, with every probe label checked
against the training data.

Our contributions are:
\begin{enumerate}
\item A controlled measurement over ten label vocabularies, two
candidate-set sizes and three probe sources, with exact paired tests,
label-level resampling and a multiple-comparison correction
(Sections~\ref{sec:design} and~\ref{sec:transfer}).
\item The finding that the release trained with label-wording augmentation is
more accurate on held-out vocabularies of ordinary and borrowed words, but not on abbreviations
or opaque codes, and that a probe at a single candidate-set size can misstate
the effect (Section~\ref{sec:transfer}).
\item Out-of-distribution and cross-model results against an open model on the
same backbone (Section~\ref{sec:unseen}).
\item Four hazards in release-to-release comparison, each with its measured
effect on numbers we had prepared for publication (Section~\ref{sec:hazards}).
\end{enumerate}

\section{Typed decisions}

A typed decision is a triple $(c, O, t)$: a context $c$, a set of candidate
options $O = \{o_1, \dots, o_k\}$, and a declared type $t \in$ \{boolean,
choice, scale\}. The model returns a distribution over $O$.

Ru-Jev\footnote{The name refers to Jev, a commercial
decision model by TypeSafe AI. Ru-Jev is an independent project, not
affiliated with TypeSafe AI and not trained on Jev outputs; Jev's API was not called in this work.} is a pointwise cross-encoder, the same family as entailment-based
zero-shot classifiers and passage rerankers. Each pair is scored independently
and the scores are normalised across options. With an encoder $f_\theta$, a
linear head $w$ and $h(c, o) = f_\theta(c, o)_{[\mathrm{CLS}]}$,
\begin{equation}
p(o_i \mid c, O) = \frac{\exp\left(w^\top h(c, o_i) / T\right)}
                        {\sum_{j=1}^{k} \exp\left(w^\top h(c, o_j) / T\right)},
\end{equation}
where $T$ is a single scalar temperature fitted after training
\citep{guo2017}, a one-parameter variant of Platt scaling \citep{platt1999}.
Training uses cross-entropy over the options of each example.

Two properties follow. The output does not depend on the order of the options,
since no option sees another. And because each option's score is computed
independently, the distribution over any subset of a candidate set is obtained
exactly by renormalising that subset's scores; we use this to evaluate the
same probe at five and at eighteen candidates from one scoring pass. The price
is cost linear in $k$. Models that read all options in one pass, such as Laya
\citep{laya}, trade order invariance for lower cost at large $k$
(Section~\ref{sec:calib}).

\section{Model and data}

The encoder is mmBERT-base \citep{marone2026mmbert}, 307M parameters of which
110M are outside the embedding matrix, released under the MIT licence. Inputs
are truncated to 256 tokens. Training uses thirteen tasks built from open
Russian-language datasets: six tasks from RussianSuperGLUE \citep{shavrina2020},
RuCoLA \citep{mikhailov2022}, one reasoning task pooling the ruOpenBookQA and
ruWorldTree sets of MERA \citep{fenogenova2024}, four classification tasks from
ruMTEB \citep{snegirev2025rumteb} (review sentiment, review rating, and the
intent and scenario tasks of MASSIVE \citep{fitzgerald2022}), and a toxicity
set \citep{dementieva2021}. The mix contains 143{,}991 training and 26{,}746
validation decisions. Each source keeps its own licence; the repository lists
them per source, including one whose README asks for research use while its
licence file is an unmodified MIT licence.

\paragraph{Preprocessing.} The data builder of v1.3 (\texttt{scripts/build\_data.py})
reduces every source row to a context, a list of options and the index of the
correct option. Texts are used as the sources give them, with no lowercasing,
normalisation, deduplication or length filtering. For RussianSuperGLUE only the
labelled train and validation splits are read and rows labelled $-1$ are
skipped; MERA questions whose answer is not one of the four letters or whose
question or an option is empty are skipped. RuCoLA was not handled correctly
(see the next paragraph). For the RussianSuperGLUE tasks the source train and validation splits are
kept, and the fields of each row are joined into a short Russian template, for
example a premise, a hypothesis and the question whether the hypothesis follows.
For the single-text sources and RuCoLA all source splits are pooled, shuffled,
capped at 30{,}000 rows, and the first 10\% form the validation split.
The pooled splits include the source test splits of review sentiment, review
rating and toxicity, with their real labels, so test-split scores on those
benchmarks would be scores on training rows; no result in this paper uses
them. For MERA
the validation split is 10\% of the labelled train split. For MASSIVE up to
12{,}000 rows of the source train split are used for training, and up to 1{,}500
rows each of the source validation and test splits for validation. Each MASSIVE example
gets the correct label and four (scenarios) or five (intents) other labels drawn
at random from the full label set. The four boolean RussianSuperGLUE tasks draw the wording of their two options
per example from a fixed list of five pairs; toxicity, review sentiment and
review rating draw one of several label sets per example
(Section~\ref{sec:aug}); RuCoLA keeps one wording. The order of the options is shuffled
in every example. The builder stops if any task has fewer than half its expected
rows (a check added for v1.3). In the current builder each shuffle that sets a split is seeded from the
task name; the builds the two releases were trained on used one shared random
stream instead (Section~\ref{sec:hazards}, hazard 1). For training and inference
the context and each option are encoded as a pair and truncated to 256 tokens,
removing tokens from the longer of the two first. Preprocessing of the
evaluation sets is described with each set in Sections~\ref{sec:design}
and~\ref{sec:unseen}.

\paragraph{Unlabelled RuCoLA test rows.}\label{sec:rucola} The labels of
RuCoLA's test split are not public, and its loader marks them $-1$. The
builder's mapping from the source label to an option index turned $-1$ into
``unacceptable'' instead of dropping the row, so both releases were trained and
validated with 2{,}788 of its 2{,}789 test sentences labelled unacceptable
(the remaining one also occurs, with its real label, in the labelled splits):
2{,}507 training and 281 validation rows for v1.1, and 2{,}536 (1.76\% of the mixture) and
252 (0.94\% of the validation split) for v1.3. We found this while preparing
this version; the released models have not been retrained, and the builder
now drops these rows.
Removing the 252 rows from the validation split of v1.3 does not change the
selected epoch, moves the temperature fitted on that split from 1.864 to 1.838
and accuracy on the full split from 0.821 to 0.824, and changes the accuracy in
each row of Table~\ref{tab:coverage} by at most 0.003. On RuCoLA itself it
matters: v1.3 scores 0.647 with these rows, against 0.599 for always answering
the majority class, and 0.689 without them, against 0.737, so on rows with a
real label it is below the majority-class rate. The figures in this paper are
those of the released model on the split it was trained with. The six
RussianSuperGLUE tasks of Table~\ref{tab:compare} contain no RuCoLA rows, so
their scores do not change when the rows are removed. A model retrained with
the v1.3 configuration on the v1.3 mixture with these rows removed, everything
else unchanged and the epoch fixed in advance, scores 0.755 on the RuCoLA rows
with a real label, against 0.689 for the release, but 0.007 lower over the
whole validation split and 0.012 lower on the six RussianSuperGLUE tasks, so
under the rule recorded before training it was not released; removing the rows
also shrinks the mixture, so the loss elsewhere cannot be attributed to the
labels rather than to the smaller mixture (\texttt{results/v131/}). The test sentences are not redistributed; their SHA-256 hashes
and the recomputed figures are in \texttt{results/rucola\_test/}.

Training runs four epochs with a checkpoint after each, and the third is
released (stored zero-indexed as epoch 2). The fourth overfits: training loss
falls to 0.085 while validation loss rises to 0.820 (Table~\ref{tab:epochs}),
and a temperature fitted on its outputs reaches 3.92, against 1.864 for the
released third epoch. Epoch selection and
temperature fitting both used the validation split, which includes the MASSIVE
validation rows; the datasets of Section~\ref{sec:unseen} played no role in
model selection.

\begin{table}[h]
\centering
\begin{tabular}{lrrr}
\toprule
Epoch & Training loss & Validation loss & Validation accuracy \\
\midrule
1 & 0.6029 & 0.5054 & 0.7536 \\
2 & 0.4336 & 0.4576 & 0.7920 \\
3 & 0.2714 & 0.4831 & \textbf{0.8181} \\
4 & 0.0849 & 0.8204 & 0.8139 \\
\bottomrule
\end{tabular}
\caption{Training trajectory of release v1.3, from the log of the training run
(\texttt{results/training/v1\_3\_train.log}); epoch checkpoints other than the
released one are not published. The third epoch is released.}
\label{tab:epochs}
\end{table}

\section{The two releases}
\label{sec:aug}

Release v1.1 presents each MASSIVE scenario label as its English key and as an
ordinary Russian word; in the rebuilt v1.1 training mix both forms occur for
all 18 scenarios, and Russian-phrase strings occur for two. Release v1.3 stores
several surface forms per label and draws one uniformly per training example.
As specified in the data builder:
\begin{itemize}
\item MASSIVE intents (60 labels): the English key, an ordinary Russian label,
the key transliterated into Cyrillic, and a clipped form of that
transliteration produced by rule;
\item MASSIVE scenarios (18 labels): the English key, an ordinary Russian
word, and a Russian phrase;
\item toxicity: five wordings of the two labels, of which v1.1 used one;
\item review sentiment: five wordings of the label set, and review rating:
four, where v1.1 used one each.
\end{itemize}
Transliterated and clipped forms are never applied to scenario labels.

The releases differ in more than the augmentation. Adding the label styles
changed the validation partitions of several tasks through a shared random
stream (Section~\ref{sec:hazards}, hazard 1); the per-task seeding that fixes
this was introduced after v1.3 was trained. The overlap audit below uses
training mixes rebuilt with the fixed builder, whose partitions differ from
those the releases were trained on; it compares option strings, which are set by
the label styles, but we did not repeat it against the exact trained builds. An
intermediate release, v1.2, changed other details. Architecture and objective are the same, but v1.1 was trained
for three epochs and v1.3 for four (with the third released), so their
learning-rate schedules differ. We
therefore attribute differences to the release, and to the augmentation only
where the probe design isolates it.

\section{Evaluation design}
\label{sec:design}

\paragraph{Questions.} All wording probes use the same 1{,}000 questions from
the source validation split of the MASSIVE scenario task (7 are duplicates).
Neither release trained on these rows, but both used them for epoch selection
and temperature fitting. Each question has one gold scenario among 18.

\paragraph{Two candidate-set sizes.} With \emph{five} candidates, each question
is shown with the gold label and four distractors drawn deterministically from
the other 17; the distractor labels are identical across all vocabularies and
models (chance 0.20). With \emph{eighteen}, every question is shown with all
labels (chance 0.056). Absolute accuracies and effect sizes are reported for
both settings and never compared across them.

\paragraph{Vocabularies.} Each vocabulary renames all 18 scenarios.
\begin{itemize}
\item \emph{Card bench}, six vocabularies written during development of the
model, before these experiments: English key, ordinary Russian word, Russian
phrase, English-derived jargon (\ru{вэзер} `weather', \ru{смолток} `small
talk', \ru{фуд-доставка} `food delivery', \ru{логистика} `logistics'),
department names (\ru{метеослужба} `weather service') and descriptive
sentences. The jargon vocabulary is not a blind test of v1.3. The intermediate
release v1.2, trained with phrase-form labels, left jargon accuracy where it was
(0.608 against 0.633 for v1.1), and the transliterated and clipped intent forms
of v1.3 were added in response, with the stated aim of moving this score; the
jargon labels themselves were never used in training. Jargon labels are also
produced by the same kind of transformation as the augmentation, Cyrillic
transliteration of an English word, here applied to scenario labels that the
transliteration rule never touches. We therefore read the jargon result as
transfer of one transformation family to a new label set.
\item \emph{Rule-derived probe}: code identifiers such as \texttt{ALARM\_SCN}
and three-letter Cyrillic abbreviations such as \ru{БУД} (for \ru{будильник}
`alarm'), generated by rule
from existing labels by the AI coding agent that trained the model, and scored
with five candidates only.
\item \emph{Independent probe}: office slang (\ru{синоптик} `forecaster',
\ru{побудка} `reveille') and opaque codes (\ru{ПГД}, \ru{БДК}, letter codes
with no stated expansion), written by a separate AI assistant session
that had not trained the model and had seen no scores for these vocabularies,
and passed on verbatim by the author.
\end{itemize}

\paragraph{Overlap with training.} Every probe label was compared with every
option string in the rebuilt training mixes of both releases, by exact equality
and by prefix in both directions, for gold labels and distractors. Two card-bench
vocabularies are training styles: the English key and the Russian word occur in
both releases' training data for all 18 labels, so they serve as controls, not
probes. The Russian phrase occurs in v1.3's training data for all 18 labels and
in v1.1's for two (one of which is identical to the Russian word), so its gain is expected by construction. The
department-name and descriptive vocabularies have no exact match in either
release, but all descriptive labels and two department labels have a prefix
relation to some training option, so neither has a strict slice; the jargon
vocabulary has three exact matches in v1.3's training data
(audio, news, weather), which its clean slice removes. Slices: for jargon, \emph{clean} drops the three labels that coincide
with the output of the transliteration rule (audio, news, weather; 15 labels,
882 questions) and \emph{strict} drops every label with an exact or prefix
relation to a training option (11 labels, 625 questions). For the independent
probe, strict drops labels with an exact or prefix relation (slang 14 labels,
codes 9);
for abbreviations, clean drops the one exact match. Every abbreviation and code
identifier has a prefix relation to some training option, so these
vocabularies have no strict slice. Filtering applies to the
gold label; distractors are not filtered, so chance on the clean and strict
slices is not exactly $1/k$. The audit uses exact equality and prefixes only; a
label that shares subwords with a training option without being a prefix of it
counts as unrelated.

\paragraph{Reading rule.} For the independent probe, a rule for reading the
eighteen-candidate slang comparison on the strict slice was recorded at 08:53
local time on 5~November~2026, after scoring had started and after the v1.1
score file had been written, but before any score file was opened. The rule
said: v1.3 higher with $p<0.05$ means the gain transfers to slang. File hashes
are in the repository. The time is stated inside the rule file, which was
first committed later the same day, so the commit history does not by itself
confirm it. We do not call this pre-registration, and the five-candidate
analysis of this probe was added afterwards. The same rule reads the opaque
codes as a lower bound; with eighteen candidates on the strict slice it
returns ``transfer'' (Section~\ref{sec:transfer}).

\paragraph{Statistics.} For each cell we report the exact McNemar test
\citep{mcnemar1947} on paired answers and a 95\% interval from a two-stage
bootstrap that resamples labels and then questions within labels
\citep{field2007}, since a label, not a question, is the unit
that generalises. With 9 to 18 labels per cell these intervals are wide and
approximate. $p$-values are Holm-adjusted \citep{holm1979} over 23 cells: all cells of
Table~\ref{tab:transfer} except the six training-style controls (English key,
Russian word and Russian phrase, in both settings), plus the two
unseen-dataset comparisons of Section~\ref{sec:unseen}. Confidence is evaluated with the area under the
risk--coverage curve (AURC, lower is better; \citealp{geifman2019}), the gap between mean confidence
and accuracy, and the number of wrong answers above 0.8 confidence per 1{,}000
questions. Ru-Jev probabilities use the released temperatures ($T=2.2754$ for
v1.1, $1.8640$ for v1.3).

\section{Which wordings the augmented release gains on}
\label{sec:transfer}

\begin{table}[t]
\centering
\resizebox{\textwidth}{!}{%
\begin{tabular}{lrrrrrrrrrr}
\toprule
& & & \multicolumn{4}{c}{Five candidates (chance 0.20)} & \multicolumn{4}{c}{Eighteen candidates (chance 0.056)} \\
\cmidrule(lr){4-7}\cmidrule(lr){8-11}
Vocabulary, slice & Labels & $n$ & v1.1 & v1.3 & $\Delta$ & 95\% CI, labels & v1.1 & v1.3 & $\Delta$ & 95\% CI, labels \\
\midrule
English key$^\dagger$ & 18 & 1000 & 0.957 & 0.962 & $+0.005$ & $[-0.01, +0.02]$ & 0.894 & 0.900 & $+0.006$ & $[-0.02, +0.03]$ \\
Russian word$^\dagger$ & 18 & 1000 & 0.956 & 0.960 & $+0.004$ & $[-0.01, +0.02]$ & 0.883 & 0.892 & $+0.009$ & $[-0.02, +0.04]$ \\
Russian phrase$^\ddagger$ & 18 & 1000 & 0.881 & 0.957 & $+0.076$ & $[+0.02, +0.14]$ & 0.703 & 0.885 & $+0.182$ & $[+0.04, +0.34]$ \\
\midrule
English-derived jargon, all & 18 & 1000 & 0.633 & 0.855 & $+0.222$$^{***}$ & $[+0.09, +0.37]$ & 0.431 & 0.640 & $+0.209$$^{***}$ & $[+0.05, +0.40]$ \\
English-derived jargon, clean & 15 & 882 & 0.624 & 0.838 & $+0.214$$^{***}$ & $[+0.08, +0.40]$ & 0.427 & 0.595 & $+0.168$$^{***}$ & $[+0.02, +0.34]$ \\
English-derived jargon, strict & 11 & 625 & 0.550 & 0.786 & $+0.235$$^{***}$ & $[+0.05, +0.45]$ & 0.328 & 0.470 & $+0.142$$^{***}$ & $[-0.04, +0.38]$ \\
Department names & 18 & 1000 & 0.835 & 0.869 & $+0.034$$^{*}$ & $[-0.03, +0.12]$ & 0.632 & 0.733 & $+0.101$$^{***}$ & $[-0.003, +0.24]$ \\
Descriptive sentences & 18 & 1000 & 0.939 & 0.958 & $+0.019$ & $[-0.001, +0.04]$ & 0.848 & 0.878 & $+0.030$$^{*}$ & $[+0.001, +0.06]$ \\
Office slang, all & 18 & 1000 & 0.648 & 0.780 & $+0.132$$^{***}$ & $[+0.01, +0.26]$ & 0.428 & 0.476 & $+0.048$ & $[-0.11, +0.21]$ \\
Office slang, strict & 14 & 773 & 0.595 & 0.761 & $+0.166$$^{***}$ & $[+0.02, +0.32]$ & 0.369 & 0.419 & $+0.050$ & $[-0.15, +0.24]$ \\
\midrule
Code identifiers & 18 & 1000 & 0.856 & 0.889 & $+0.033$$^{*}$ & $[-0.01, +0.09]$ & \multicolumn{4}{c}{not scored} \\
Abbreviations, all & 18 & 1000 & 0.403 & 0.379 & $-0.024$ & $[-0.18, +0.17]$ & \multicolumn{4}{c}{not scored} \\
Abbreviations, clean & 17 & 959 & 0.420 & 0.360 & $-0.060$$^{*}$ & $[-0.20, +0.12]$ & \multicolumn{4}{c}{not scored} \\
Opaque codes, all & 18 & 1000 & 0.265 & 0.270 & $+0.005$ & $[-0.08, +0.09]$ & 0.104 & 0.146 & $+0.042$$^{**}$ & $[-0.01, +0.10]$ \\
Opaque codes, strict & 9 & 512 & 0.135 & 0.168 & $+0.033$ & $[-0.04, +0.13]$ & 0.018 & 0.061 & $+0.043$$^{**}$ & $[+0.01, +0.09]$ \\
\bottomrule
\end{tabular}}
\caption{Accuracy of the two releases on the same 1{,}000 questions, varying
only label wording. $\Delta$ is v1.3 minus v1.1. Stars give Holm-adjusted
exact McNemar $p$ over the 23 non-control cells: $^{*}<0.05$, $^{**}<0.01$,
$^{***}<0.001$. The label-level interval resamples labels, then questions.
$^\dagger$Training style for both releases (control). $^\ddagger$Training style
for v1.3 only. The rule-derived probe (code identifiers, abbreviations) was
scored with five candidates only. Raw McNemar counts and per-label results are
in the repository.}
\label{tab:transfer}
\end{table}

\paragraph{Ordinary and borrowed words.} On every held-out vocabulary made of
words, the augmented release is more accurate in both settings
(Table~\ref{tab:transfer}). The largest gains are on English-derived jargon:
$+0.235$ with five candidates and $+0.142$ with eighteen on the strict slice,
which excludes every label with an exact or prefix relation to a training
option. Office slang gains $+0.166$ with five candidates (label-level interval
above zero) and $+0.050$ with eighteen; the recorded reading rule for the
latter returns ``transfer'' (raw $p=0.016$), but the gain does not survive the
Holm correction and its label-level interval spans zero. Department names gain
$+0.034$ and $+0.101$, descriptive sentences $+0.019$ and $+0.030$. The
direction is the same in all fourteen word-vocabulary cells (the all, clean and
strict slices of one vocabulary are nested, and both candidate-set sizes come
from one scoring pass, so these cells are not independent evidence); the size depends on the
candidate-set size in different ways for different vocabularies, so a probe at
one size alone can misstate the effect. On the strict slice with eighteen candidates the slang result
is a reshuffle as much as a lift: 5 labels improve and 6 get worse, some by
half the scale (\ru{справочное бюро} `information desk' 0.000 to 0.585, \ru{перечни}
`lists' 0.534 to 0.086).

\paragraph{One label behind several gains.} Part of these gains is a lean
rather than reading. The scenario \texttt{qa} (106 of the 1{,}000 questions) is
a label the earlier release almost never chooses when it is worded as jargon,
a department name or slang (not so for abbreviations or codes): with
eighteen candidates v1.1 answers it 7, 21 and 4 times on jargon, department
names and slang, against 73, 113 and 79 times for v1.3. That one label supplies
62 of the net 48 questions on slang with eighteen candidates, so without it the
difference turns to $-0.016$, and 36 of the net 34 on department names with
five candidates ($-0.002$ without it). Fitting one bias per option with the
temperature on the 1{,}000 questions of each vocabulary, out of fold as in
Section~\ref{sec:unseen}, separates the lean from reading. The jargon gain
largely survives ($+0.182$ of $+0.222$ with five candidates, $+0.217$ of
$+0.209$ with eighteen, $+0.145$ on the strict slice with five). Slang with
five candidates keeps about half ($+0.066$ of $+0.132$), department names keep
$+0.044$ and $+0.069$, and slang with eighteen candidates keeps $+0.064$ on the
strict slice over questions ($+0.032$ to $+0.097$), but its label-level interval
($-0.04$ to $+0.17$) spans zero. Slices are read after a fit on the whole
vocabulary, because a fit on a slice can push options that never occur on it
towards minus infinity, which a caller cannot do. Per-label contributions and
the fits are in \texttt{results/vocab\_bias/}.

\paragraph{Abbreviations and codes.} On three-letter abbreviations v1.3 is not
better. On the 17 labels unrelated to training it is lower as released,
$-0.060$ (Holm $p=0.017$), but 52 of the 58 net questions are the one label
\texttt{qa} (0.689 to 0.198); without it the difference is $-0.007$, v1.3 is
better on 10 of the 17 labels and worse on 6, and after the per-option
correction the difference is $0.000$ ($-0.030$ to $+0.031$ over questions). The
near tie on all 18 labels ($-0.024$) comes from the one label
that occurs in training, on which v1.3 gains 0.829. Opaque codes are close to
unreadable for both releases once overlapping labels are removed. With eighteen
candidates on the strict slice v1.3 gains $+0.043$ (Holm $p=0.002$, label
interval above zero), so by the recorded reading rule the gain transfers, but
it moves v1.3 from below chance (v1.1, 0.018) to barely above it (0.061 against
0.056), which we do not count as usable, and after the per-option correction
it is $+0.008$ on the strict slice and $-0.024$ on all labels; with five candidates neither release
differs from the other. Code identifiers such as
\texttt{ALARM\_SCN} contain the trained key and are easy for both, so we do not
treat them as a test of robustness.

\paragraph{Confidence.} In most cells where accuracy improves, the risk--coverage
area falls, so a caller who thresholds on confidence is better served (among
non-control cells, the only exception is slang on all labels with eighteen candidates, 0.354 to 0.395): on strict jargon
AURC goes from 0.235 to 0.078 with five candidates and from 0.452 to 0.418 with
eighteen; on strict slang from 0.223 to 0.121 with five candidates and is
unchanged (0.476 against 0.470) with eighteen. The count of confident errors
tells a less favourable story. With five candidates it falls on jargon (110 to
62 per 1{,}000 on the strict slice). With eighteen it rises on jargon (79 to
109 on all labels) and on slang (151 to 191 on the strict slice), consistent with the
augmented release being more confident overall, while it falls on department
names (123 to 87). On abbreviations it rises (77 to
118 on the clean slice). The augmented release is therefore not, by itself, safer
to threshold on unfamiliar wording; a deployment should refit the
temperature on its own labels.

\paragraph{Reading.} The release trained on several surface forms per label is
more accurate on unfamiliar labels that are words, ordinary or borrowed,
including office slang, a kind of wording absent from training, although the
slang gain is partly the lean described above. Two runs of
the v1.3 schedule (three checkpoints) that, through the error described in
Section~\ref{sec:hazards}, were trained with the phrase forms but without the
transliterated forms reached 0.588 to 0.619 on jargon, the v1.1 level, and
0.879 to 0.885 on department names, at or above the v1.3 level (five
candidates, raw scores, two seeds). Their mixture was at the v1.2 level, with
all of v1.2's wording augmentation and partitions of its own, so this
accidental partial ablation is consistent with the jargon gain coming from the
transliterated forms and the department-name gain from the earlier wording
augmentation, without isolating either. It gives no usable
gain on abbreviations or opaque codes, where the label string carries little
of its meaning. Because the releases differ in more than the augmentation
(Section~\ref{sec:aug}) and each was trained once, these are differences
between two trained models, consistent with the augmentation as their cause
but not a controlled ablation of it.

\section{Unseen datasets and an open comparison model}
\label{sec:unseen}

Two Russian datasets were used by neither release and are not named in Laya's
model card, which does not list its training data: news headline topics from ruMTEB (six classes; licence declared
as MIT in the repository metadata, no licence file) and single-label rows of
CEDR emotions \citep{sboev2021cedr} (five classes; Apache-2.0 in the metadata).
Laya \citep{laya} is a 322M model on the same mmBERT backbone that reads all
options in one pass. It was given a one-line instruction, which its interface
requires and Ru-Jev does not take; option keys were neutral, so the option text
carried the meaning for both.

\begin{table}[h]
\centering
\small
\begin{tabular}{lrrrrrr}
\toprule
& \multicolumn{3}{c}{News headlines ($k=6$)} & \multicolumn{3}{c}{Emotions ($k=5$)} \\
\cmidrule(lr){2-4}\cmidrule(lr){5-7}
Model & Acc. & Conf.$-$acc. & AURC & Acc. & Conf.$-$acc. & AURC \\
\midrule
Ru-Jev v1.1 & \textbf{0.652} & $+0.028$ & 0.209 & 0.455 & $+0.085$ & 0.444 \\
Ru-Jev v1.3 & 0.568 & $+0.095$ & 0.260 & 0.511 & $-0.064$ & 0.302 \\
Laya        & 0.610 & $+0.242$ & 0.221 & \textbf{0.605} & $+0.056$ & 0.256 \\
\bottomrule
\end{tabular}
\caption{Datasets neither Ru-Jev release trained on, 1{,}000 questions each. Ru-Jev uses its
released temperatures; Laya returns its own probabilities.}
\label{tab:unseen}
\end{table}

No model is more accurate on both datasets (Table~\ref{tab:unseen}). v1.3 is
8.4 points below v1.1 on headlines (Holm $p<10^{-6}$; five of six labels worse,
none better) and 5.6 points above it on emotions. Laya is markedly
overconfident on headlines and not on emotions. When each model refits its own
temperature on half the rows and is scored on the other half, residual
calibration error is 0.056 to 0.067 for Ru-Jev and 0.055 and 0.077 for Laya
(25 splits with the recorded seed for both). We make no claim
of superiority over Laya in accuracy or calibration.

\paragraph{A label prior accounts for most of the headline gap.} A temperature divides every
score by the same number and cannot change which option is chosen. Most of the
headline drop is a shift in label priors, chiefly towards \ru{политика}
`politics': v1.3 answers \ru{политика} 318
times for 166 political headlines (v1.1: 232), its recall on \ru{экономика} `economy'
falls from 0.730 to 0.448, and 41 of the 153 headlines that only v1.1 gets
right are economy headlines read as politics. We therefore fit one additive
bias per option jointly with the temperature,
$p(o_i \mid c, O) \propto \exp\left(s_i / T + b_{o_i}\right)$, on one half of
the rows and score the other half, over 25 random splits with both halves
scored (Table~\ref{tab:bias}). On headlines the biases lift v1.3 from 0.568 to
0.640 and v1.1 from 0.652 to 0.673, and the paired difference shrinks from
$-0.084$ to $-0.034$ (95\% bootstrap interval $-0.060$ to $-0.006$ over
questions and $-0.067$ to $+0.001$ over labels). On emotions they lift v1.1
from 0.455 to 0.556 and v1.3 from 0.511 to 0.571, and the difference of
$+0.014$ is not distinguishable from zero over questions or over labels. On
headlines the remaining gap is still below zero over questions, so part of the
drop is not a prior: after the fit, recall on \ru{политика} is 0.703 for v1.1
and 0.617 for v1.3 and on \ru{экономика} 0.734 and 0.655, so these two labels
carry about $-2.8$ of the $-3.4$ points. Before the fit both releases have the
same recall on \ru{политика} (0.771), while v1.3 sends 71 economy headlines to
politics against 35 for v1.1. What remains looks like a boundary between two
labels that v1.3 draws worse and that the per-option biases do not repair here;
after the fit, v1.3's additional errors on political headlines go to economy
and to incidents. Neither training mix contains a news-topic task. Scoring the
intermediate checkpoints (\texttt{results/unseen/ladder.md}) separates two
effects. The shift of economy headlines towards \ru{политика} appears with
v1.2 and in the first epoch of v1.3, while the ranking of the two labels,
measured by the AUC of the score margin between \ru{экономика} and
\ru{политика} on the 340 headlines of these two labels, which no temperature or bias can
change, stays at the v1.1 level (0.928) through v1.2 (0.929) and the first two
epochs of v1.3 (0.921, 0.938) and falls only at the released third epoch
(0.895) and the fourth (0.875). A second seed of the v1.1 recipe gives 0.909,
and the three checkpoints trained without the transliterated forms give 0.834
to 0.921, so the loss of separation lies within the variation between seeds
and is not attributable to the transliterated forms. The biases also absorb each dataset's class balance, so
these are adapted results on 500 labelled rows, not zero-shot ones. The
intervals are not Holm-adjusted, the bootstrap keeps each row's out-of-fold
result fixed rather than refitting the biases, so it understates the
variability of the fit, and with six and five labels the label-level intervals
are unreliable. Biases belong to the label set they are fitted on; the released
calibration script fits them on the same labelled rows as the temperature. This
analysis was suggested by a reader after the preprint was posted and was run to
test that suggestion.

\begin{table}[h]
\centering
\small
\begin{tabular}{lrrrr}
\toprule
& \multicolumn{2}{c}{News headlines ($k=6$)} & \multicolumn{2}{c}{Emotions ($k=5$)} \\
\cmidrule(lr){2-3}\cmidrule(lr){4-5}
Model & Released & With biases & Released & With biases \\
\midrule
Ru-Jev v1.1 & 0.652 & 0.673 & 0.455 & 0.556 \\
Ru-Jev v1.3 & 0.568 & 0.640 & 0.511 & 0.571 \\
\midrule
v1.3 $-$ v1.1 & $-0.084$ & $-0.034$ & $+0.056$ & $+0.014$ \\
\bottomrule
\end{tabular}
\caption{Accuracy on the unseen datasets with the released temperatures and
with one bias per option fitted jointly with the temperature on half of the
rows and scored on the other half (25 random splits, both halves scored).
Differences are paired over questions.}
\label{tab:bias}
\end{table}

\paragraph{A third unseen dataset and the choice of checkpoint.} After the
biases, the ten distinct checkpoints we kept (the two releases, v1.2, the four
epochs of v1.3 (the third is the release), a second seed of the v1.1 recipe and three checkpoints of the
two stale-mixture runs) rank differently on headlines and on emotions (Spearman
$-0.17$; $-0.16$ as released), and the second of the four epochs of v1.3, the
best on headlines after the fit, is not ahead of the released third on emotions ($-0.008$, interval
spanning zero). To test whether either ranking holds elsewhere we added
scientific-rubric classification (ru-SciBench, GRNTI rubrics; 990 test texts of
ten rubrics, five candidates, no text overlap with training) and recorded the
reading rule before any row was scored: a ranking holds if its Spearman
correlation with the new set after the biases is at least 0.5. Neither holds
($-0.18$ against headlines, $-0.41$ against emotions), and each of the three
sets picks a different best checkpoint, so no single unseen set should choose
the checkpoint for another. On this set v1.3 is ahead of v1.1 after the biases
by $+0.091$ ($+0.059$ to $+0.125$ over questions, $+0.010$ to $+0.184$ over
labels) and ahead of the second v1.1 seed by $+0.045$. All 28 rubrics have the
same test size, so the ten were chosen by a tie-break; after the verdict we
scored all 28 and drew twenty random sets of ten. In none of them does a
Spearman correlation reach 0.5, and v1.3 stays ahead of the second seed in
every draw ($+0.037$ to $+0.101$, both intervals above zero in all twenty),
and when the four rubrics whose names share a prefix with a training option
are excluded the median is $+0.066$, with the interval over questions above zero
in all twenty draws and the one over labels in nineteen. On the tie-break ten
alone, without those four rubrics, the lead is $+0.026$ and its interval spans
zero. These intervals price questions and labels, not the training seed: over
the same twenty draws the two runs of the v1.1 recipe differ by a median of
$+0.070$, and the lead of v1.3 over the second v1.1 seed is no larger than that
in 11 of the 20. The figures are in \texttt{results/unseen/ladder\_bias.md} and
\texttt{results/unseen3/}.

\paragraph{A second run of the v1.3 recipe.}\label{sec:seed2} The runs on the
v1.3 schedule described in Section~\ref{sec:transfer} trained on a stale
mixture without the transliterated and clipped intent forms, so the released
v1.3 had no valid second seed. We trained one: the v1.3 configuration with seed
1337 on the mixture rebuilt with the builder of the release (198 intent surface
forms, identical row counts), with the epoch fixed in advance at the third of
four and the reading rule recorded before training. The rule: the gap between
the recipes, the mean of the two v1.3 runs minus the mean of the two v1.1 runs
after per-option biases, is established on a dataset if it exceeds both
within-recipe spreads. On jargon it is, by a wide margin: the gap is $+0.182$
with eighteen candidates and $+0.159$ with five, against spreads of at most
0.074, and at a fixed seed the transliterated and clipped forms alone are worth
$+0.262$ and $+0.201$ (the new run against the stale-mixture run with the same
seed). On the scientific rubrics it is established only narrowly: over the
twenty draws the median gap is $+0.078$ against spreads of $-0.039$ for v1.3 and
$+0.070$ for v1.1, and the rule holds in 11 of the 20; there the forms are worth
$-0.007$ at a fixed seed, so what separates the recipes on this dataset is not
the forms. On headlines ($-0.022$) and emotions ($-0.039$) the gap is not
established, and the two v1.3 runs differ by 0.086 on emotions. The fixed
epoch is not the new run's best on those two datasets (its first epoch reads
0.702 and 0.590 against 0.626 and 0.484), which agrees with the checkpoint
result above, and on jargon with eighteen candidates the new run is 0.074
below the released v1.3, so the release is the stronger of the two v1.3 runs
there. The figures are in \texttt{results/v13\_seed2/}.

On the card bench Laya scores well below Ru-Jev (with eighteen candidates,
0.150 against 0.640 on jargon). The MASSIVE training split is in Ru-Jev's mix,
and Laya's own card reports near-chance zero-shot performance on typed
decisions, so that gap measures home advantage and is not reported as a
comparison.

\section{In-distribution accuracy, calibration and cost}
\label{sec:calib}

\paragraph{In-distribution accuracy.} On the six RussianSuperGLUE tasks, whose
rows and option sets are identical in both builds, the releases answer exactly
8{,}120 of 10{,}157 decisions correctly and disagree on 1{,}680, 840 in each
direction (McNemar $p=1.0$; paired standard error of the difference 0.0040).
Per-row predictions and the SHA-256 of both weight files are in the
repository. The word-sense task supplies 84\% of these decisions.

\begin{table}[h]
\centering
\small
\begin{tabular}{lrrrrr}
\toprule
Task & v1.1 & v1.3 & Only v1.1 right & Only v1.3 right & $n$ \\
\midrule
russe (word sense)      & 0.822 & 0.824 & 650 & 667 & 8505 \\
danetqa (yes/no)        & 0.742 & 0.753 & 58 & 67 & 821 \\
terra (entailment)      & 0.775 & 0.733 & 39 & 26 & 307 \\
parus (causal choice)   & 0.680 & 0.590 & 22 & 13 & 100 \\
rcb (three-way inference)     & 0.450 & 0.509 & 23 & 36 & 220 \\
rwsd (coreference)      & 0.564 & 0.480 & 48 & 31 & 204 \\
\midrule
All six                 & 0.7994 & 0.7994 & 840 & 840 & 10157 \\
\bottomrule
\end{tabular}
\caption{Identical rows and option wording for both releases.}
\label{tab:compare}
\end{table}

\paragraph{Calibration.} Over 25 random half-splits of these rows (seed
recorded in the repository), fitting a temperature on one half and scoring the
other, v1.3 takes a lower temperature in every split (1.50--1.60 against
1.94--2.10): its raw scores are less overconfident. Residual expected
calibration error \citep{naeini2015} after each release receives its own
temperature is the same (0.0190 against 0.0191; the sign of the difference
changes across splits).

\paragraph{Selective prediction.} With a calibrated probability, a caller can
answer above a threshold and route the rest to a person \citep{elyaniv2010}.
In the system for citizen appeals that motivated this work, every prediction
was reviewed by an operator and that review, not the model, dominated handling
time \citep{beskorovainyi2026appeals}; a threshold is the natural next step.
Table~\ref{tab:coverage} gives the trade-off for v1.3 on its own validation
split, with the temperature fitted on the disjoint half. Because the same split
informed epoch selection, these figures are optimistic, and
Sections~\ref{sec:transfer} and~\ref{sec:unseen} show that the released
temperature does not carry over to unfamiliar labels or new domains. A script
in the repository refits it on a few hundred labelled rows.

\begin{table}[h]
\centering
\begin{tabular}{rrrr}
\toprule
Share answered & Threshold & Accuracy on answered & Errors per 1000 \\
\midrule
50\%  & 0.85 & 0.967 & 33 \\
70\%  & 0.69 & 0.921 & 79 \\
90\%  & 0.54 & 0.859 & 141 \\
100\% & 0.23 & 0.823 & 177 \\
\bottomrule
\end{tabular}
\caption{Coverage against accuracy, v1.3, on the 13{,}373 validation
decisions held out from temperature fitting. Accuracy on this half is 0.823; the
same scoring pass gives 0.821 on the full validation split of 26{,}746
decisions (the trainer's own log, Table~\ref{tab:epochs}, gives 0.818).}
\label{tab:coverage}
\end{table}

\paragraph{Cost.} Table~\ref{tab:latency} gives the median latency of one
decision on one Apple M4 CPU using all ten cores, over 25 runs. Ru-Jev is 1.2 times slower than Laya
with two options and 2.4 times slower with eighteen.

\begin{table}[h]
\centering
\begin{tabular}{lrrr}
\toprule
Options $k$ & 2 & 5 & 18 \\
\midrule
Ru-Jev v1.3 (ms) & 18.5 & 32.8 & 73.8 \\
Laya (ms)        & 15.5 & 17.8 & 30.5 \\
\bottomrule
\end{tabular}
\caption{Median latency per decision, Apple M4 CPU, all ten cores, 25 runs after 5 warm-up runs.}
\label{tab:latency}
\end{table}

\section{Four hazards in comparing releases}
\label{sec:hazards}

Our first comparison of the releases reported a clean accuracy gain and two
large regressions. None of the three was real. We list the hazards because none
is exotic and each was invisible in the output of a comparison that did not
test for it.

\paragraph{1. Splits coupled through a shared random stream.} The data builder
carved each task's validation slice with the module-level random generator,
seeded once and consumed by every other random call in the file. Adding label
styles to one task changed the number of draws consumed before the next task
was carved, and every later task received a different partition. Intersecting
the two builds shows the result is leakage: 90.3\% of the new acceptability
validation rows and 89.4\% of the reasoning rows had been in the v1.1 training
data, along with 47.2\% of review rating and 21.6\% of toxicity. The apparent
regression on acceptability (0.909 to 0.647) was v1.1 scored on its own
training rows; on held-out rows it scores 0.650. All three figures are on
validation splits that contain the unlabelled RuCoLA rows of
Section~\ref{sec:rucola} (\emph{Unlabelled RuCoLA test rows}); without them v1.3 scores 0.689 against a
majority-class rate of 0.737, and v1.1 has not been rescored. The fix seeds each carve from
the task's identity, \verb|random.Random(f"split:{task}")|. We verified it by
rebuilding the previous release's data from its commit and confirming that the
validation rows of the six tasks whose option sets never changed were identical
to the earlier build.

\paragraph{2. Matching rows on input text alone.} Matching validation rows by
input text gave nine comparable tasks. Under the augmented recipe a row with
the same text can carry an option wording the baseline never saw, so scoring
the baseline there measures robustness, not accuracy. Requiring identical text
\emph{and} option wording leaves six tasks. A toxicity gain of 0.722 to 0.954
that we had presented as a capability gain was of this kind: v1.3 trained on
all five toxicity wordings, v1.1 on one, and v1.1 scores 0.961 with the wording
it trained on.

\paragraph{3. Mixed denominators.} A summary row computed on one random half of
the comparable rows was printed beside a per-task table counting all
10{,}157, and a small calibration advantage for v1.3 was read off that one
split. On all rows accuracy is identical (Table~\ref{tab:compare}), and the
calibration difference changes sign across splits. Both figures had reached the
published model card and were withdrawn. The original output is kept in the
repository; the half-split figures cannot be regenerated from the released
per-row scores, so we do not quote them. A related
slip appeared in our drafts: a worst-case gap computed as the best vocabulary
on all labels minus the jargon vocabulary on the clean slice.

\paragraph{4. Integrity checks against a stale reference.} To estimate seed
variance we retrained both recipes on a remote machine and checked its training
mix against a local copy by SHA-256, row count and per-task counts. All
matched. The local copy predated the commit that added the transliteration
styles, so the seed runs meant to replicate v1.3 were trained without them;
their jargon accuracy (0.588 to 0.619 across runs and checkpoints) fell near
v1.1's. Row counts could not
catch this, since the new styles change option strings, not the number of rows.
Counting distinct option strings per task did (120 for intents in the stale
copy, 198 after rebuilding; some generated forms coincide). These runs are
kept in the repository as a record; they enter no release comparison and are
used only as the partial ablation of Section~\ref{sec:transfer}.

\paragraph{Protocol.} Seed every partition from the identity of what is
partitioned; define comparability over everything the model consumes and print
the overlap; measure leakage directly; choose a control that a change must not
move; keep one denominator per table; verify artefacts by content that the
change is expected to alter, not by hashes against a copy of unknown age; and
evaluate wording robustness at more than one candidate-set size.

\section{Limitations}

\emph{Few training runs per recipe.} Each recipe now has two runs on the
unseen datasets and on the full vocabularies (the second v1.3 run is described
under ``A second run of the v1.3 recipe'' in Section~\ref{sec:unseen}), and the
v1.1 recipe a third on the card bench, which bounds the seed noise without estimating it. On the card bench the two
further v1.1 runs on jargon with five candidates score 0.658 and 0.622 against 0.633 for the release (a
range of 0.036; 0.046 on the clean slice), but on the strict slice, where the
largest gain is reported, they score 0.648 and 0.563 against 0.550, a range of
0.098, and two runs of the same recipe differ there with exact McNemar
$p=4\times10^{-7}$. The released v1.1 is the weakest of the three on that slice,
so the raw $+0.235$ is $+0.138$ against the strongest run. After the per-option
correction the three runs lie within 0.041 and v1.3 is ahead by $+0.145$, about
three and a half times that range. On the other bench vocabularies the range is
0.004 to 0.021, and the differences on the English-key and Russian-word
controls lie within it. The second v1.3 run was scored on full vocabularies
only, so the strict-slice figure rests on one v1.3 run, the noise in a difference between two recipes is
larger than in one recipe, and two or three runs per recipe give a range, not
a distribution.

\emph{Training data.} Both releases were trained with about 2{,}500 RuCoLA test
sentences under a label they do not have (Section~\ref{sec:rucola},
\emph{Unlabelled RuCoLA test rows}). Removing them from the validation split
changes the other reported accuracies by at most 0.003 and the fitted
temperature by 0.026. What the released weights would score had they been
trained without them is not known: a retraining without them scored 0.007 lower
on the validation split and 0.012 lower on the six RussianSuperGLUE tasks,
cannot separate the label from the smaller mixture, and was not released.

\emph{Not a single-factor comparison.} The releases differ in data partitions
and in toxicity wording as well as in label augmentation
(Section~\ref{sec:aug}). The released v1.3 is the best of four checkpoints
by validation accuracy, while v1.1 is the last of three, and both selected the
epoch and fitted the temperature on a split that contains the MASSIVE rows used
by the wording probes. The out-of-distribution changes in particular cannot be
attributed to the augmentation alone.

\emph{Vocabularies.} Each style is one vocabulary of 18 labels, so style and
vocabulary are confounded, and label-level intervals rest on 9 to 18 clusters.
All vocabularies were written by the author's project or by AI agents working
in it; none was collected from deployed systems. The independent probe was
written by a model of the same family as the coding agent that trained
Ru-Jev, and that session had seen the card-bench and rule-derived results but no
score for its own vocabularies, so its independence is procedural. Abbreviations and code
identifiers were scored with five candidates only.

\emph{Scope.} Wording probes cover one routing task in Russian. The model
answers only within the options it is given and cannot report that none fits;
a none-of-these option must be supplied explicitly. One temperature serves all
tasks and does not carry over to new labels or domains; per-option biases are
specific to the label set they are fitted on and need labelled rows from it. Reasoning tasks with
fewer than a thousand training rows remain weak; a 560M backbone
(xlm-roberta-large) tried once did not improve them.

\section{Conclusion}

When the caller supplies the label set, an evaluation that varies the model but
not the label wording measures the easy case. Varying the wording at two
candidate-set sizes showed that the release trained on several surface forms per label
is more accurate on unfamiliar labels that are words, ordinary or borrowed, and
gives no usable gain on abbreviations or opaque codes, where its apparent loss
on abbreviations is one label that a per-option bias repairs; and that better accuracy on unfamiliar
wording does not by itself bring fewer confident errors when the candidate set
is large. A second run of each recipe keeps the jargon gain far above seed
noise. A practical reading is
to choose augmentation styles that resemble the wording a deployment will use,
to refit the temperature, and per-option biases where the model leans towards
one option, on that wording, and to evaluate on styles absent from
the augmentation at more than one candidate-set size. Because rankings of
checkpoints disagree across unseen datasets, an epoch or a seed should not be
chosen on one of them. We also recommend the
release comparison protocol of Section~\ref{sec:hazards}.

\section*{Data and code availability}

{\sloppy Weights of both releases are on Hugging Face
(\url{https://huggingface.co/besktech/ru-jev}; v1.1 on a branch). Training
code, the data builder, all label vocabularies, the overlap audit, the latency
benchmark, per-row scores for every experiment in this paper, the analysis
script that produces Table~\ref{tab:transfer} including AURC and confident
errors, the SHA-256 of both weight files and of the files referred to in the
reading rule, and the recorded split seed are at
\url{https://github.com/beskvladimir-create/ru-jev}; the release used for
this paper (v1.3.0) is archived at Zenodo,
\href{https://doi.org/10.5281/zenodo.23986847}{doi:10.5281/zenodo.23986847}.
The per-option bias analysis and the checkpoint ladder of
Section~\ref{sec:unseen}, the training log and an English codebook for all
result files are in release v1.3.6,
\href{https://doi.org/10.5281/zenodo.24075942}{doi:10.5281/zenodo.24075942},
where \texttt{results/analysis/unseen\_bias.py} reproduces
Table~\ref{tab:bias} and \texttt{results/analysis/unseen\_ladder.py} the
checkpoint figures. The per-option correction on the probe vocabularies is in
v1.3.8, \href{https://doi.org/10.5281/zenodo.24125472}{doi:10.5281/zenodo.24125472};
the checkpoint rankings, the scientific-rubric dataset with its reading rule and
the full score matrix are in v1.3.9,
\href{https://doi.org/10.5281/zenodo.24152275}{doi:10.5281/zenodo.24152275};
the second run of the v1.3 recipe is in v1.3.11,
\href{https://doi.org/10.5281/zenodo.24185900}{doi:10.5281/zenodo.24185900};
the account of the unlabelled RuCoLA rows, the corrected builder and a
retraining without these rows are in v1.3.13,
\href{https://doi.org/10.5281/zenodo.24218670}{doi:10.5281/zenodo.24218670}; later additions are in the
latest release,
\href{https://doi.org/10.5281/zenodo.23986846}{doi:10.5281/zenodo.23986846}.
Code and weights are
under the MIT licence; training data keep their own licences, listed per
source in the repository.\par}

\section*{Acknowledgements}

Claude Code and Claude (Anthropic) were used as tools in this work: to write
and run the experiment code, to generate two probe vocabularies, and in
preparing the text. The author designed the study, made the decisions and
takes full responsibility for the contents.

We thank the Hugging Face user dipankarsarkar, who pointed out the label prior
behind the headline result and proposed the per-option biases of
Section~\ref{sec:unseen}, and whose further questions led to the checkpoint,
third-dataset and second-seed analyses.

\end{document}